%% file: report.tex
\documentclass[10pt,twocolumn,letterpaper]{article}

\usepackage{cvpr}
\usepackage{times}
\usepackage{epsfig}
\usepackage{graphicx}
\usepackage{amsmath, amsthm, amssymb}
\usepackage{mdwlist}

\usepackage[ruled,vlined]{algorithm2e}

\usepackage[pagebackref=true,breaklinks=true,letterpaper=true,colorlinks,bookmarks=false]{hyperref}

\theoremstyle{definition}		

\newcommand{\squishlist}{
 \begin{list}{$\bullet$}
  { \setlength{\itemsep}{0pt}
     \setlength{\parsep}{3pt}
     \setlength{\topsep}{3pt}
     \setlength{\partopsep}{0pt}
     \setlength{\leftmargin}{1.5em}
     \setlength{\labelwidth}{1em}
     \setlength{\labelsep}{0.5em} } }

\newcommand{\squishend}{
  \end{list}  }

\newcommand{\figwidth}{6.85in}

\newcommand{\ARXIV}{}

\cvprfinalcopy 

\def\cvprPaperID{XXX} 

\begin{document}

\title{Template Adaptation for Face Verification and Identification}

\author{
Nate Crosswhite, Jeffrey Byrne \\
Systems \& Technology Research (STR) \\
{\tt\small \{nate.crosswhite, jeffrey.byrne\}@stresearch.com}
\and
Chris Stauffer  \\
Visionary Systems and Research (VSR) \\
{\tt\small stauffer@vsandr.com}
\and
Omkar M. Parkhi, Qiong Cao, Andrew Zisserman  \\
University of Oxford \\
{\tt\small \{omkar, qiong, az\}@robots.ox.ac.uk}
}

\maketitle

\input{template_adaptation}

{\small
\bibliographystyle{ieee}
\bibliography{janus}
}

\end{document}

%% file: template_adaptation.tex
\begin{abstract}
Face recognition performance evaluation has traditionally focused on one-to-one verification, popularized by the Labeled Faces in the Wild dataset \cite{Huang07} for imagery and the YouTubeFaces dataset \cite{Wolf11} for videos. In contrast, the newly released IJB-A face recognition dataset \cite{Klare15} unifies evaluation of one-to-many face identification with one-to-one face verification over {\em templates}, or sets of imagery and videos for a subject.  In this paper, we study the problem of template adaptation, a form of transfer learning to the set of media in a template.  Extensive performance evaluations on IJB-A show a surprising result, that perhaps the simplest method of template adaptation, combining deep convolutional network features with template specific linear SVMs, outperforms the state-of-the-art by a wide margin.  We study the effects of template size, negative set construction and classifier fusion on performance, then compare template adaptation to convolutional networks with metric learning, 2D and 3D alignment.  Our unexpected conclusion is that these other methods, when combined with template adaptation, all achieve nearly the same top performance on IJB-A for template-based face verification and identification.  
\end{abstract}



\section{Introduction}

Face recognition performance using deep learning has seen dramatic improvements in recent years.  Convolutional networks trained with large datasets of millions of images of thousands of subjects have shown remarkable capability of learning facial representations that are invariant to age, pose, illumination and expression (A-PIE) \cite{Parkhi15,Taigman14,Schroff15,Taigman15,Sun14,Sun15}.  These representations have shown strong performance for recognition of imagery and video in-the-wild in unconstrained datasets, with recent approaches demonstrating capabilities that exceed human performance on the well known Labeled Faces in the Wild dataset \cite{Huang07}.  

The problem of face recognition may be described in terms of face verification and face identification.  Face verification involves computing a one-to-one similarity between a probe image and a reference image, to determine if two image observations are of the same subject.  In contrast, face identification involves computing a one-to-many similarity between a probe media and a gallery of known subjects in order to determine a probe identity. Face verification is important for access control or re-identification tasks, and face identification is important for watch-list surveillance or forensic search tasks.  

Face recognition performance evaluations have traditionally focused on the problem of face verification.  Over the past fifteen years, face datasets have steadily increased in size in terms of number of subjects and images, as well as complexity in terms controlled vs. uncontrolled collection and amount of A-PIE variability \cite{Miller15}.   The Labeled Faces in the Wild dataset \cite{Huang07} contains 13233 images of 1680 subjects, and compares specific pairs of images of subjects to characterize 1:1 verification performance.  Similarly, the YouTubeFaces dataset \cite{Wolf11} contains 3425 videos of 1595 subjects, and compares pairs of videos of subjects for verification.  These datasets have set the established standard for face recognition research, with steadily increasing performance \cite{Lu15,Taigman14,Schroff15,Parkhi15}.  Recently, protocols for face identification have been introduced for LFW \cite{Rowden14} to address the performance evaluation for identification on a common dataset.  However, the imagery in LFW was constructed with a well known near-frontal selection bias, which means evaluations are not predictive of performance for large in-the-wild pose variation.  In fact, recent studies have shown that while algorithm performance for near frontal recognition is equal to or better than humans, performance of automated systems at the extremes of illumination and pose are still well behind human performance \cite{Phillips15}.  

The IJB-A dataset \cite{Klare15} was created to provide the newest and most challenging dataset for both verification and identification.  This dataset includes both imagery and video of subjects manually annotated with facial bounding boxes to avoid the near frontal bias, along with protocols for evaluation of both verification and identification.  Furthermore, this dataset performs evaluations over {\em templates} \cite{Grother14} as the smallest unit of representation, instead of image-to-image or video-to-video.  A template is a set of all media (images and/or videos) of a subject that are to be combined into a single representation suitable for matching.  Template based representations are important for many face recognition tasks, which take advantage of an historical record of observations to further improve performance. For example, a template provides a useful abstraction to capture the mugshot history of a criminal for forensic search in law enforcement, or lifetime enrollment images for visa or driver's licenses in civil identity credentialing for improved access control.  Biometric templates have been studied for face recognition, where performance on older algorithms have increased given an historical set of images \cite{Grother14}. The IJB-A dataset is the only public dataset that enables a controlled evaluation of template-based verification and identification at the extremes of pose, illumination and expression.

In this paper, we study the problem of {\em template adaptation}.  Template adaptation is an example of transfer learning, where the target domain is defined by the set of media of a subject in a template.  In general, transfer learning includes a source domain for feature encoding of subjects trained offline, and a specific target domain with limited available observations of new subjects.  In the case of template adaptation, the source domain may be a deep convolutional network trained offline to predict subject identity, and the target domain is the set of media in templates of never before seen subjects.  In this paper, we study perhaps the simplest form of template adaptation based on deep convolutional networks and one-vs-rest linear SVMs.  We combine deep CNN features trained offline to predict subject identity, with a simple linear SVM classifier trained at test time using all media in a template as positive features to classify each new subject.  



Extensive evaluation of template adaptation on the IJB-A dataset has generated surprising results.  First, template adaptation outperforms all top performing techniques in the literature: convolutional networks combined with triplet loss similarity \cite{Schroff15,Parkhi15,Sankaranarayanan16}, joint Bayesian metric learning \cite{Chen16}, pose specialized networks \cite{AbdAlmageed16},  2D alignment \cite{Parkhi15}, 3D frontalization \cite{Taigman14} and novel convolutional network architectures \cite{RoyChowdry16}.  Second, template adaptation when combined with these other techniques results in nearly equivalent performance.  Third, we show a clear tradeoff between the size of a template (e.g. the number of unique media in the template) and performance, which leads to the conclusion that if the average largest template size is big enough, then a simple template adaptation strategy is the best choice for both verification and identification on template based datasets.

\ifdefined\ARXIV
\input{fig_cartoon}
\fi

\section{Related Work}

The top performing approaches for face verification on Labeled Faces in the Wild \cite{Huang07} and YouTubeFaces \cite{Wolf11} are all based on convolutional networks.  VGG-Face is the application of the VGG-16 convolutional network architecture \cite{Simonyan15} trained on a newly curated dataset of 2.6M images of 2622 subjects.  This representation includes triplet loss embedding and 2D alignment for normalization to provide state of the art performance.  
FaceNet \cite{Schroff15} applied the inception CNN architecture \cite{Szegedy15} to the problem of face verification.  This approach included metric learning to train a triplet loss embedding to learn a 128 dimensional embedding optimized for verification and clustering.  This network was trained using a private dataset of over 200M subjects.  DeepFace \cite{Taigman14}\cite{Taigman15} uses a deep network coupled with 3D alignment, to normalize facial pose by warping facial landmarks to a canonical position prior to encoding.  DeepID2+ \cite{Sun15} and DeepID3 \cite{Sun14} extended the inception architecture to include joint Bayesian metric learning \cite{Chen12} and multi-task learning for both identification and verification.  

These top performing convolutional network architectures have interesting common properties. First, they all exhibit deep convolutional network structure, often with parallel specialized sub-networks. However, Parkhi et. al \cite{Parkhi15} showed that a the VGG-16 very deep architecture  \cite{Simonyan15}, when trained with a broad and deep dataset containing one thousand examples of 2622 subjects, outperformed networks with specialized networks \cite{Schroff15} and  ensembles \cite{Sun14} on YouTubeFaces.  Second, many top performing approaches use some form of pose normalization such as 2D/3D alignment \cite{Taigman14,Parkhi15,AbdAlmageed16} to warp the facial landmarks into a canonical frontal pose.  Finally, many approaches use metric learning in the form of triplet loss similarity or joint Bayesian metric learning for the final loss to learn an optimal embedding for verification \cite{Schroff15,Parkhi15,Chen16}.  An recent independent study reached a similar conclusion that multiple networks combined in an ensemble and metric learning are crucial for strong performance on LFW \cite{Hu15}.

Recent evaluations on IJB-A \cite{Klare15} are also based on convolutional networks and mirror the top performing approaches on LFW and YouTubeFaces.  Recent approaches include deep networks using triplet loss similarity\cite{Sankaranarayanan16}\cite{Chen15} and joint Bayesian metric learning \cite{Chen16}, and five pose specialized sub-networks with 3D pose rendering \cite{AbdAlmageed16}.  Face-BCNN \cite{RoyChowdry16} applies the bilinear CNN architecture to face identification, publishing the earliest results on IJB-A.

Transfer learning has been well studied in the literature, and we refer to a comprehensive survey on the topic \cite{pan2010survey}.  
Transfer learning and domain adaptation for convolutional networks is typically performed by pretraining the network on a labeled source domain, replacing the final loss layer for a new task, then fine-tuning the network on this new objective using data from the target domain \cite{Razavian14}.  Prior work has shown that freezing the network, then replacing a softmax loss layer with a linear SVM loss can result in improved performance for classification tasks \cite{Tang13,Huang06}.  These approaches can be further improved by jointly training the SVM loss and the CNN parameters, so that the lower level features are fine-tuned with respect to the SVM objective \cite{Tang13}.  However, such retraining requires a large target domain training set to fine tune all parameters in the deep network.  In this paper, we focus on updating the linear SVM only, as this classifier has a regularization structure that has been shown to perform well for unbalanced training sets with few positive examples (e.g. from media in a template) and many negative examples \cite{Malisiewicz11}\cite{Kobayashi15}.  

Finally, we note that the approach of defining a similarity function for face verification using linear SVMs trained on a large negative set was originally proposed as one-shot similarity (OSS) \cite{Wolf09}\cite{Wolf11b}.  We study the more general form of this original idea, by considering templates of images and videos of varying size, alternative fusion strategies, and the impact of gallery negative sets for identification.

\ifdefined\ECCV
\input{fig_cartoon}
\fi

\section{Template Adaptation}
\label{s:template_adaptation}

Template adaptation is a form of transfer learning, combining deep convolutional network features trained on a source domain of many labeled faces, with template specific linear SVMs trained on a target domain using the media in a template.  Template adaptation can be further decomposed into {\em probe adaptation} for face verification and {\em gallery adaptation} for face identification.
In this section, we describe these approaches.

First, we provide preliminary definitions.  A media observation $x$ is either a color image of a subject, or a set of $m$ video frames of a subject.      
An image encoding $z=f(x)$ is a mapping $f(x) \in \mathcal{R}^d$ from an image $x$ to an encoding $z$ with dimensionality $d$ (e.g. features from a deep CNN).  An average encoding $\bar{z} = \frac{1}{m} \sum_x f(x)$ is the average of image/frame encodings in a media observation, such as the encodings for all frames in a video.  A template $X$ is a set of encoded media observations $X = \{f(x_1), f(x_2),\ldots, f(x_k)\}$ of one subject.  The size of a template $|X|$ is defined as the number of unique media used for encoding.  Finally, a gallery $G=\{(X_1,y_1),(X_2,y_2),\ldots,(X_m,y_m)\}$ is a set of tuples of templates $X$ and associated subject identification label $y$. 

Figure \ref{f:cartoon} shows an overview of this concept.  Each colored shape corresponds to a feature encoding of image or a video feature for the media in a template, such as generated from a convolutional network trained offline. The gray squares correspond to encodings of a large set of media of unique subjects that are very likely to be disjoint from any template.  The centroid of the colored shapes corresponds to the average encoding for this template.  
Probe adaptation is the problem of max-margin classification of the positive features from a template to the large negative feature set.  The similarity between the blue probe template and the mated (genuine subject) green template is the margin (dotted lines) of the green feature encodings to the decision surface.  Observe that this margin is positive, whereas the margin for the red classifier is negative, so that the blue/green similarity is much larger than blue/red as desired.  Gallery adaptation is the problem of max-margin classification where the negative feature set for the gallery templates are defined by the other gallery templates.  Observe that adding the magenta subject causes the decision surface for the red and green classifiers to shift, improving the margin score for the probe.

More formally, probe adaptation is the training of a similarity function $s(P,Q)$ for a probe template $P$ and reference template $Q$.  Train a linear SVM for $P$, using unit normalized average encodings of media in $P$ as positive features and a large feature set as negatives. The large negative set contains one feature encoding for many subject identities, so this set is very likely to be disjoint with the probe template. 
Similarly, train a linear SVM for $Q$, using the unit normalized average encodings for media in $Q$ as positive features and a large feature set as negatives.  Finally, let $P(q)$ be notation for evaluating the SVM functional margin (e.g. $w^Tx$) trained on $P$, and evaluated using the unit normalized average media encoding $q$ in template $Q$.  The final similarity score for probe adaptation is the fusion of the two classifier margins using a linear combination $s(P,Q) = \frac{1}{2}P(q) + \frac{1}{2}Q(p)$.  For implementation details, see section \ref{ss:experimental_system}.

Gallery adaptation is the training of a similarity function $s(P,G)$ from a probe template $P$ to gallery $G$.  A gallery contains templates $G=\{X_1,X_2,\ldots,X_m\}$, and gallery adaptation trains a linear SVM for all pairs $s(P,X_i)$ following the approach for probe adaptation.  Gallery adaptation differs from probe adaptation in that the large negative set for a template $X_i$ is all unit normalized media encodings from all other templates in $G$ not including $X_i$.  In other words, the other non-mated subjects in the gallery are used to construct negative features for $X_i$, whereas the large negative set is used for $P$.  The final similarity score for gallery adaptation is the fusion of the probe classifier and the gallery classifier for each $X \in G$ using the linear combination $s(P,X) = \frac{1}{2}P(x) + \frac{1}{2}X(p)$.







\section{Results}
\label{s:results}

The proposed approach in section \ref{s:template_adaptation} introduces a number of research questions to study.  
\medskip

\noindent {\bf How does this compare to the state of the art?}  In section \ref{ss:ijba_evaluation}, we compare the template adaptation approach to all published results and show that the proposed approach exceeds the state of the art by a wide margin.  Furthermore, in section \ref{ss:aoa} we perform an analysis of alternatives to combine the state of the art techniques with template adaptation and show that when combined, these alternative approaches all result in nearly the same performance.
\smallskip

\noindent {\bf How should the negative set be formed?}  Template adaptation requires training linear SVMs, which require a labeled set of positive and negative feature encodings.  In section \ref{ss:negativeset}, we perform a study to evaluate different strategies of constructing this negative set including using a holdout set, external negative set and combinations. Results show that the gallery based negative set is best for gallery adaptation, and a holdout set derived from the same dataset as the templates is best for verification.  
\smallskip

\noindent {\bf How large do the templates need to be?}  In section \ref{ss:templatesize}, we study the effect of template size, or total number of media in a template, on verification performance to identify the minimum template size necessary, to help guide future template based dataset construction.  We show that a minimum of three unique media per template results in diminishing returns for template adaptation.  
\smallskip

\noindent {\bf How should template classifier scores be fused?}  In section \ref{ss:fusion}, we study the effect of different strategies for combination of two classifiers, based on winner take all and weighted combinations of on template size. We conclude that an average combination is best with winner take all a close second.

\noindent {\bf What are the error modes of the template adaptation?}  In section \ref{ss:errors}, we visualize the best and worst templates pairs in IJB-A for verification (identification errors are shown in the supplementary material), and we show that template size (e.g. number of media in a template) has the largest effect on performance.  
\smallskip

\ifdefined\ARXIV
\input{fig_ijba_evaluation_arxiv}
\fi

\subsection{Experimental System}
\label{ss:experimental_system}
We use the VGG-Face deep convolutional neural network \cite{Parkhi15}, using the penultimate layer output as the feature encoding $f$.  
For computing the average encoding across frames of video, we use {\em face tracks} which compute the mean encoding of all frames in a video followed by unit normalization.  This approach was shown to be effective for Fisher vector encoding  \cite{Parkhi14} and 
deep CNN encoding \cite{Parkhi15}. 

Media encoding is preprocessed according to the following pipeline.  
For each media, we crop each face using the ground truth or detected facial bounding box dilated by a factor of 1.1.  Then, we anisotropically rescale this face crop to 224x224x3, such that the aspect ratio is not preserved.  This is the assumed input size for the CNN.  Next, we encode this face crop for each image or frame in the template using the VGG-face network, and compute average video encodings for each video.  Next, we unit normalize each media feature, and train the weights and bias for a linear SVM for each template.  We use the LIBLINEAR library with L2-regularized L2-loss primal SVM with class weighted hinge loss objective  \cite{Fan08}.  
\ifdefined\ARXIV  
\begin{equation}
\begin{split}
\min_{w} \frac{1}{2}w^Tw + C_p \sum_{i=1}^{N_p} \mathtt{max} [0, 1-y_iw^Tx_i]^2 + \\ C_{n} \sum_{j=1}^{N_{n}} \mathtt{max} [0, 1-y_j w^Tx_j]^2  
\label{e:loss}
\end{split}
\end{equation}
\else  
\begin{equation}
\min_{w} \frac{1}{2}w^Tw + C_p \sum_{i=1}^{N_p} \mathtt{max} [0, 1-y_iw^Tx_i]^2 + C_{n} \sum_{j=1}^{N_{n}} \mathtt{max} [0, 1-y_j w^Tx_j]^2  
\label{e:loss}
\end{equation}
\fi

The loss in (\ref{e:loss}) includes terms for both positive and negative features, such that $C_p$ is the regularization constant for $N_p$ positive observations ($y_i = +1$) and $C_n$ for negative observations ($y_i=-1$).  This formulation of the loss enables data rebalancing for cases where $N_p << N_n$.  The positive features in $N_p$ are the average media encodings in the template.  The negative features are derived from a large negative feature set in $N_n$ (either from a large negative set for probe adaptation, or other non-mated templates for gallery adaptation).  The parameters $C_p=C\frac{N_p+N_n}{2N_p}$ and $C_n=C\frac{N_p+N_n}{2N_n}$ adjust the regularization constants to be proportional to the inverse class frequency.  The parameter $C=10$ in the SVM, trading-off regularizer and loss, was determined using an held-off validation subset of the data.  Finally, the learned weights $w$ include a bias term by augmenting $x$ with a constant dimension of one. 

At test time, we evaluate the linear SVMs as described in section \ref{s:template_adaptation}.  We compute the average media encodings for each media in a template, then compute the mean of the media encodings, then unit normalize forming a template encoding. This constructs a single feature for each template.  Given two templates $P$ and $Q$, let the notation $P(q)$ be the evaluation of the functional SVM margin (e.g. $P(x)=w^Tx$) for the trained linear SVM for $P$, given the template encoding $q$ for $Q$.  Finally, the similarity $s(P,Q)=\frac{1}{2}P(q) + \frac{1}{2}Q(p)$ is a weighted combination of the functional margins for the SVM for $P$ evaluated on template encoding $q$ and $Q$ evaluated on $p$.

For baseline comparison, we use the VGG-face network with the output of the 4096d features from the penultimate fully connected layer layer.  Media encodings are constructed by averaging features across a video \cite{Parkhi14,Parkhi15}, and template encodings are constructed by averaging media encodings over a template, then unit normalizing.  Template similarity is equivalent to negative L2 distance over unit normalized template encodings.  We also compare results with 2D alignment, triplet similarity embedding and joint Bayesian triplet similarity embedding.  For the triplet loss and joint Bayesian metric learning, we use hyperparameter settings such that minibatch = 1800, 1M “semi-hard” \cite{Schroff15} negative triplets per minibatch, dropconnect regularization \cite{Wan13}, 3 epochs of Parallel SGD \cite{Zinkevich11}, fixed learning rate $\nu=0.25$.  For 2D alignment, we use ground truth facial bounding boxes and facial landmark regression \cite{Kazemi14}, followed by a robust least squares similarity transform estimation to a reference box to best center the nose.  

For all research studies in sections \ref{ss:aoa} - \ref{ss:errors}, we report 1:1 verification ROC curve for all probe and gallery template pairs in IJB-A split 1 and CMC for identification on IJB-A split 1 (see section \ref{ss:ijba_evaluation} for definitions).   This is equivalent to IARPA Janus Challenge Set 2 (CS2) evaluation protocol, which is also reported in the literature.

\subsection{IJB-A Evaluation}
\label{ss:ijba_evaluation}

\ifdefined\ECCV
\input{fig_ijba_evaluation}
\fixme_update_figure_to_match_arxiv
\fi


In this section, we describe the results for evaluation of the experimental system on the IJB-A verification and identification protocols \cite{Klare15}. IJB-A contains 5712 images and 2085 videos of 500 subjects, for an average of 11.4 images and 4.2 videos per subject.  This dataset was manually curated using Mechanical Turk from media-in-the-wild to annotate the facial bounding box and eyes and nose facial landmarks, and this manual annotation avoids the Viola-Jones near-frontal bias. Furthermore, this dataset was curated to control for ethnicity, country of origin and pose biases.

Metrics for 1:1 verification are evaluated using a decision error tradeoff (DET) curve.  The 1:1 DET curve is equivalent to a receiver operating characteristics (ROC) curve, where the true accept rate is one minus the false negative match rate.   
This evaluation plots the false negative match rate vs. the false match rate as a function of similarity threshold for a given set of pairs of templates for verification.

Metrics for 1:N identification are the Decision Error Tradeoff (DET) curve and the Cumulative Match Characteristic (CMC) curve.  The 1:N DET curve plots the false negative identification rate vs. the false positive identification rate as a function of similarity threshold for a search of L=20 candidate identities in a gallery.  The 1:N CMC curve is an information retrieval metric that captures the recall of a specific probe identify within the top-K most similar candidates when searching the gallery.  
This DET curve is appropriate for limiting the workload for an analyst by allowing for a similarity threshold to be applied to reject false matches even if in the top-K.  For detailed description of these metrics, refer to \cite{Klare15,Grother14}.

\ifdefined\ARXIV
\input{fig_negativeset}
\fi

Performance evaluation for IJB-A requires evaluation of ten random splits of the dataset into training and testing (gallery and probe) sets. The evaluation protocol for 1:1 verification considers specific pairs of mated (genuine) and non-mated (imposter) subjects.  The non-mated pairs were chosen to control for gender and skin tone to make the verification problem more challenging.  Performance is reported for operating points on each of the curves:  1:1 DET reports false negative match rate at a false match rate of 1e-2, 1:N DET report true positive identification rate (e.g. 1-false negative identification rate) at false positive identification rate of 1e-2, and CMC report true positive identification rate (recall or correct retrieval rate) at rank-one and rank-ten.  The 10 splits are used to compute standard deviations for each of these operating points, to characterize statistical significance of the results. 

Figure \ref{f:ijba_evaluation} shows the overall evaluation results on IJB-A.  This evaluation compares the baseline approach of VGG-Face only \cite{Parkhi15} with the proposed approach of VGG-Face encoding with probe and gallery template adaptation.  These results show that identification performance is slightly improved for rank 1 and rank 10 retrieval, however there are large performance improvements for the 1:N DET for identification and the 1:1 DET for verification.  The table in figure \ref{f:ijba_evaluation} shows performance at specific operating points for verification and identification, and compares to published results in the literature for joint Bayesian metric learning \cite{Chen16}, triplet similarity embedding \cite{Sankaranarayanan16}, multi-pose learning \cite{AbdAlmageed16}, bilinear CNNs \cite{RoyChowdry16} and very deep CNNs \cite{Parkhi15,Wang15}.  These results show that the proposed template adaptation, while conceptually simple, exhibits state-of-the-art performance by a wide margin on this dataset.

\subsection{Analysis of Alternatives}
\label{ss:aoa}

\ifdefined\ECCV
\input{fig_analysis_of_alternatives}
\fi
\ifdefined\ARXIV
\input{fig_analysis_of_alternatives}
\fi

Figure \ref{f:aoa} shows an analysis of alternatives study.  The state of the art approaches on LFW and YouTubeFaces often augment a very deep CNN encoding with metric learning \cite{Schroff15,Parkhi15} for improved verification scores or 2D alignment \cite{Taigman14,Parkhi15} to better align facial bounding boxes.  In this study, we implement triplet loss similarity embedding, joint Bayesian similarity embedding and 2D alignment, and use these alternative feature encodings as input to template adaptation.  In this study, we seek to answer whether these alternative strategies will provide improved performance over using CNN encoding only or CNN encoding with template adaptation.

We report 1:1 DET for all probe and gallery  template pairs in IJB-A split 1 and CMC for identification on IJB-A split 1.  This study shows that template adaptation on the CNN output provides nearly the same result as template adaptation with metric learning or 2D alignment based features.  This implies that the additional training and computational requirements for these approaches are not necessary for template based datasets.  Furthermore, this study shows that 2D alignment does not provide much benefit on IJB-A, in contrast with reported performance on near frontal datasets \cite{Parkhi15,Taigman14}.  One hypothesis is that this is due to the fact that this dataset has many profile faces for which facial landmark alignment is inaccurate or fails altogether.

\ifdefined\ARXIV
\input{fig_templatesize}

\fi

\subsection{Negative Set Study}
\label{ss:negativeset}

Figure \ref{f:negativeset} shows a negative set study.  We study the effect of different combinations of negative feature sets on overall verification performance.  Recall that probe and gallery template adaptation require the use of a large negative set for training each linear SVM.  This study compares using combinations of features drawn from the non-mated subjects in the gallery (neg) and features drawn from an independent subject disjoint training set (trn). This training set is drawn from the same dataset distribution as the gallery, but is subject disjoint.  

The results in figure \ref{f:negativeset} show that using the gallery set as negative feature set provides the best performance for gallery adaptation.  Using the disjoint training set for probe adaptation is the best for verification.  This is the final strategy used for evaluation in figure \ref{f:ijba_evaluation}.  This conclusion is somewhat surprising that the probe adaptation was worse when constructing a negative set combining neg+trn, as a larger negative set typically results in better generalization performance for related approaches such as  exemplar-SVM \cite{Malisiewicz11}.  However, a larger negative set would dilute the effect of the discriminating between gallery subjects, which is the primary goal of the evaluation, so a focused negative set would be appropriate.

\ifdefined\ECCV
\input{fig_negativeset}
\fi

Next, we experimented with the CASIA Web-Face dataset \cite{Yi14}.  The best negative set for probe adaptation is a set drawn from the same distribution as the templates.  However, in many operatational conditions, this dataset will not be available. 
To study these effects, 
we constructed a dataset by sampling 70K images from CASIA balanced over classes, and pre-encoding these images for template adaptation training.  Figure \ref{f:negativeset} (bottom) shows that this results in slightly reduced verification performance.  One hypothesis is that this imagery exhibits an unmodeled dataset bias for IJB-A faces, or that CASIA is image only, while IJB-A is imagery and videos. 

\ifdefined\ECCV
\input{fig_templatesize}
\fi

%



\subsection{Template Size Study}
\label{ss:templatesize}

Figure \ref{f:template} shows an analysis of performance as a function of template size.  For this study, we consider pairs of templates $(P,Q)$ and compute the maximum template size as $\max(|P|,|Q|)$.   Next, we consider max template sizes in the range $(1,2), (2,4), (4,8), (8,16), (16,32) and (32,64)$, and compute a verification ROC curve for only those template pairs with sizes within the range. For each, we report a single point on the ROC curve at a false alarm rate of 1e-2 or 1e-3.  Results from section \ref{ss:ijba_evaluation} show that the largest benefit for template adaptation is on verification performance, so we analyze the effect of the template sizes on this metric.   

Figure \ref{f:template} (left) shows mean similarity score for templates of mated subjects only within a given template size range.  This shows that as the template size increases the mated similarity score also increases.  This is perhaps not surprising, as the more observations of media that are available in a template, the better the subject representation and the better the similarity score.  The largest uncertainty as shown by the error bars is when the maximum template size is one, which is also not too surprising.  Interestingly, the variance on the similarity scores does not decrease as template sizes increase, rather they stay largely the same even as the mean similarity increases.

Figure \ref{f:template} (right) shows the effect of template size on verification performance.  For each point on this curve, we split the dataset into templates that contained sizes within the range shown.  Then, we computed a ROC curve and report the true match rate at a false alarm rate of 1e-3 and 1e-2, an operating point on the verification ROC curve.  This result shows that the rate of increase in performance is largest for few media, and performance saturates at about 3 media per template.  Furthermore, as the number of media per template increases, the verification score at 1e-2 increases by about 19\% from one media per template to sixty four.  This also shows that the largest benefit for template adaptation is when there are at least three media per template.

\subsection{Fusion Study}
\label{ss:fusion}

Figure \ref{f:fusion} shows a study for comparing three alternatives for fusion of classifiers.  Recall from section \ref{ss:experimental_system} that a final similarity score is computed as a linear combination of the classifiers trained for templates $P$ and $Q$.  In this section, we study different strategies for setting this weighting. 

In general, the template classifier fusion from section \ref{s:template_adaptation} is a linear combination of SVM functional margins,  $s(P,Q)=\alpha P(q) + (1-\alpha)Q(p)$.  We explore strategies based on winner take all ($\alpha \in \{0,1 \}$), template weighted fusion ($\alpha=|P|/(|P|+|Q|)$ and an experiment using the SVM geometric margin (e.g. $P(x) = w^Tx / |w|$), as suggested in \cite{Kobayashi15}.  The default strategy is average fusion such that $\alpha=0.5$.  Results show that the strategy of computing a weighted average with $\alpha=0.5$ of probe and gallery templates is the best strategy.  We also performed a hyperparameter search over $\alpha$, which confirmed this selection.  

Finally, we also note that we ran experiments computing average media encodings, computing the margins for each encoding, then averaging the margins.  This strategy performed consistently worse than computing average feature encodings.

\subsection{Error Analysis}
\label{ss:errors}

Finally, we visualized identification and verification errors in different performance domains, in order to gain insight into template-based facial recognition.  This analysis provides a better understanding of the error modes to better inform future template-based facial recognition.  More detailed figures and additional discussion, including identification analysis, are available in supplemental material.

Figure \ref{f:verification_errors} shows four columns of verification probe and gallery pairs for: the best scoring mated pairs; worst scoring mated pairs; best scoring non-mated pairs; and worst scoring non-mated pairs.  After computing the similarity for all pairs of probe and gallery templates, we sort the resulting list.  Each row represents a probe and gallery template pair.  The templates contain from one to dozens of media.  Up to eight individual media are shown with the last space showing a mosaic of the remaining media in the template.  Between the templates are the IJB-A Template IDs for probe and gallery as well as the best mated and best non-mated scores.

Figure \ref{f:verification_errors} (far left) shows the highest mated similarities.  In the thirty highest scoring correct matches, we immediately note that every gallery template contains dozens of media. The probe templates either contain dozens of media or one media that matches well.  Figure \ref{f:verification_errors} (center left) shows the lowest mated template pairs, representing failed identification.  The thirty lowest mated similarities result from single-media probe templates that are low contrast, low resolution, extremely non-frontal, or not oriented upwards.
 
Figure \ref{f:verification_errors} (center right) showing the worst non-mated pairs highlights very understandable errors involving single-media probe templates representing impostors in challenging orientations.  Figure \ref{f:verification_errors} (far right) showing the best non-mated similarities shows the most certain non-mates, again often involving large templates.

\ifdefined\ARXIV
\input{fig_fusion}

\fi
\ifdefined\ECCV
\input{fig_fusion}
\fi

\section{Conclusions}

In this paper, we have introduced template adaptation, a simple and surprisingly effective strategy for face verification and identification that achieves state of the art performance on the IJB-A dataset.  Furthermore, we showed that this strategy can be applied to existing networks to improve performance.  Futhermore, our evaluation provides compelling evidence that there are many face recognition tasks that can benefit from a historical record of media to aid in matching, and that this is an important problem to further evaluate with new template-based face datasets.  


Our analysis shows that performance is highly dependent on the number of media available in a template.  This strategy results in performance that results in 19\% decrease in verification scores when a template contains a single media, such as comparing image to image or video to video, as in LFW or YouTubeFaces style evaluations.  However, when probe or gallery templates are rich and at least one template contains greater than three media, performance quickly saturates and dominates the state of the art.  

\ifdefined\ECCV
\input{fig_cs2_d21_verification}

\fi

Finally, it remains to be seen if the conclusions hold for other datasets.  The IJB-A dataset is currently the only public dataset with a template based evaluation protocol, and it may be that our performance claims are due to dataset bias, even though the composition of this dataset was engineered to avoid systemic bias \cite{Klare15}.  Finally, the gallery size for this dataset is limited to 500 subjects, and it remains to be seen if the performance claims scale as the number of subjects increase.


\smallskip 
\noindent {\bf Acknowledgment.}
%
%
This research is based upon work supported by the Office of the Director of National Intelligence (ODNI), Intelligence Advanced Research Projects Activity (IARPA) under contract number 2014-14071600010. The views and conclusions contained herein are those of the authors and should not be interpreted as necessarily representing the official policies or endorsements, either expressed or implied, of ODNI, IARPA, or the U.S. Government.  The U.S. Government is authorized to reproduce and distribute reprints for Governmental purpose notwithstanding any copyright annotation thereon.

\ifdefined\ARXIV
\input{fig_cs2_d21_verification}
\fi

%% file: fig_cartoon.tex
\begin{figure*}[!t]
\begin{centering}
\includegraphics[width=\figwidth{}]{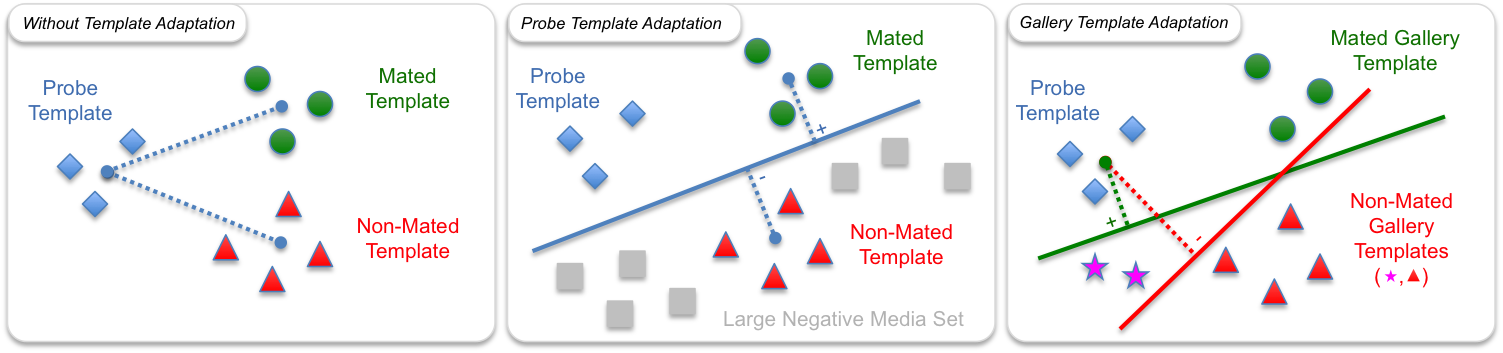} 
\caption{Template Adaptation Overview.  (left) Without template adaptation, the probe template is about equally similar as shown by the dotted lines to the mated and non-mated templates.  (middle) With probe adaptation, a max-margin classifier separates the probe template features from a large negative feature set, which increases the mated template similarity and decreases the non-mated.
 (right) With gallery adaptation, a max-margin classifier separates each gallery template features from all other gallery templates, which results in desired decrease in similarity between the probe and non-mated template.  
}
\label{f:cartoon}
\end{centering}
\end{figure*}

%% file: fig_ijba_evaluation_arxiv.tex
\begin{figure*}[t]
\begin{centering}
\includegraphics[width=\figwidth{}]{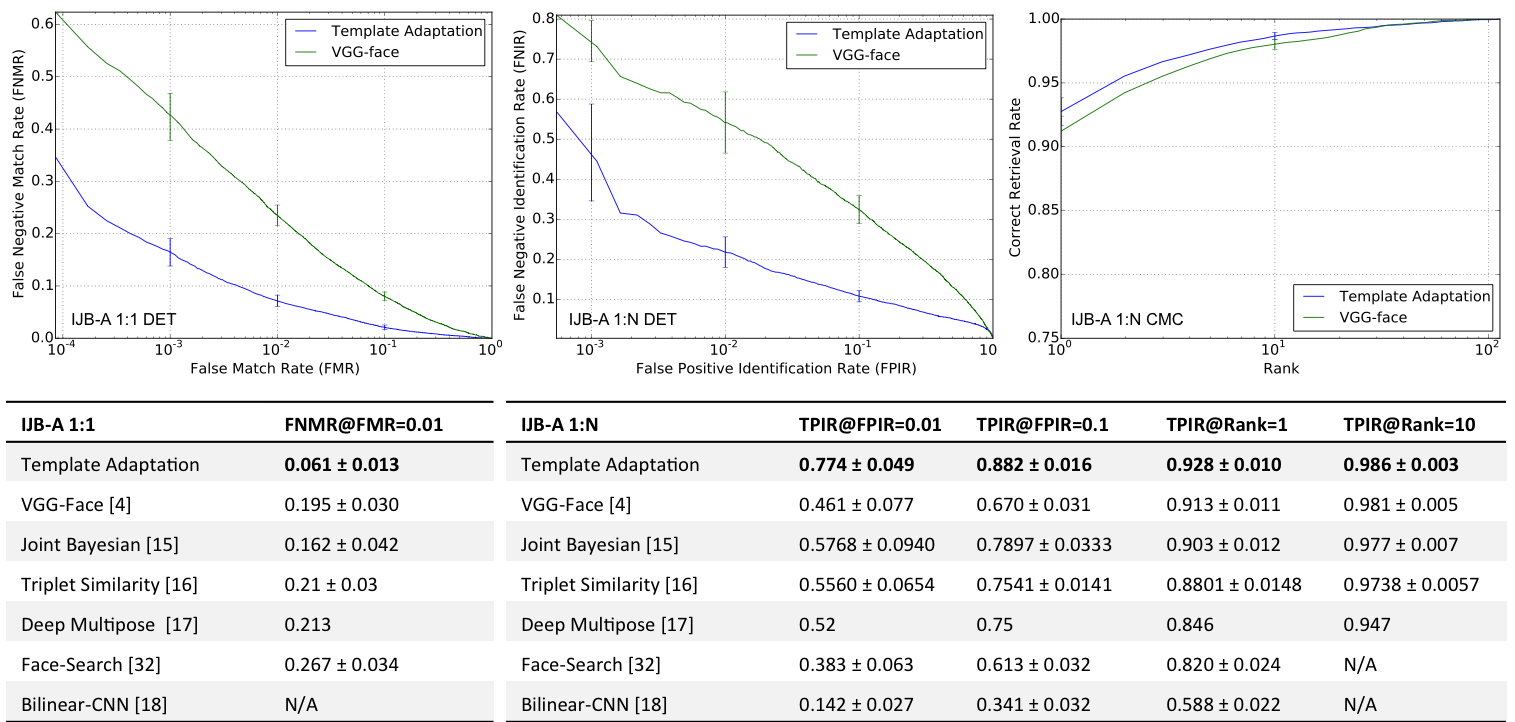} 
\caption{IJB-A Evaluation.  (top) 1:1 DET for verification, 1:N DET for identification and CMC for identification shown for template adaptation and VGG-face \cite{Parkhi15}.  (bottom) Performance at operating points as compared to published results sorted by rank-1 recall (true positive identification rate or TPIR) for VGG-face \cite{Parkhi15}, Bilinear-CNN \cite{RoyChowdry16}, Joint Bayesian \cite{Sankaranarayanan16}, Triplet Similarity \cite{Chen16}, Face-Search \cite{Wang15} and Deep Multipose \cite{AbdAlmageed16}.  Results show that Template Adaptation sets a new state-of-the-art by a wide margin.}
\label{f:ijba_evaluation}
\end{centering}
\end{figure*}

%% file: fig_negativeset.tex
\begin{figure*}[!t]
\begin{centering}
\includegraphics[width=\figwidth{}]{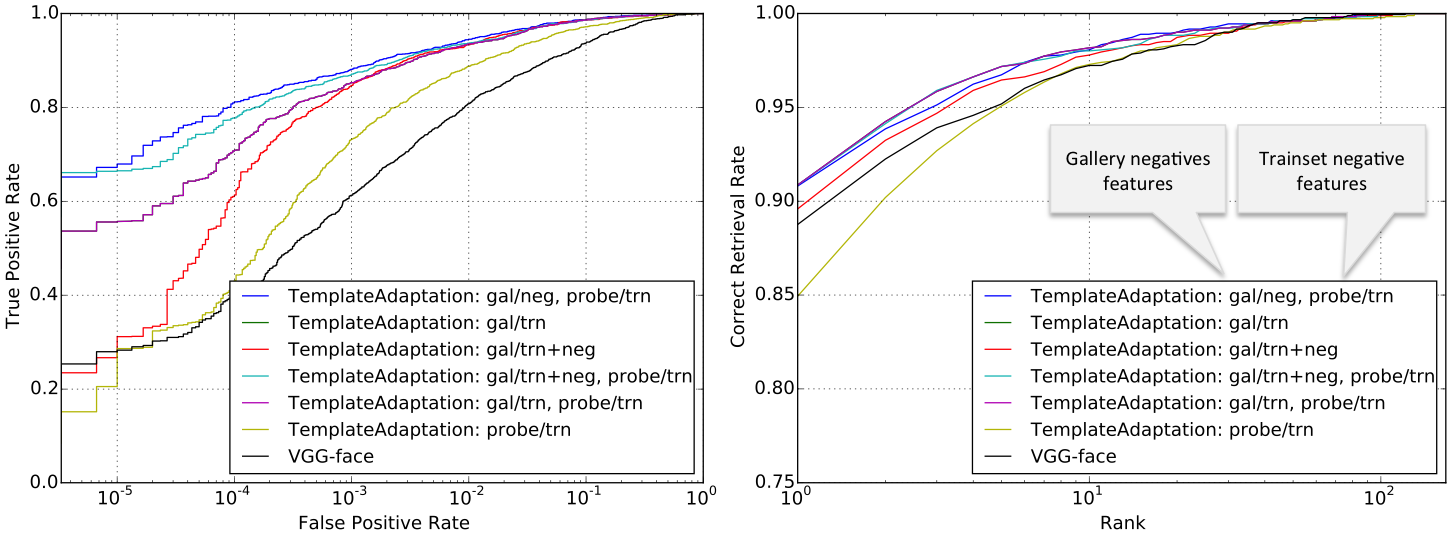} 
\includegraphics[width=\figwidth{}]{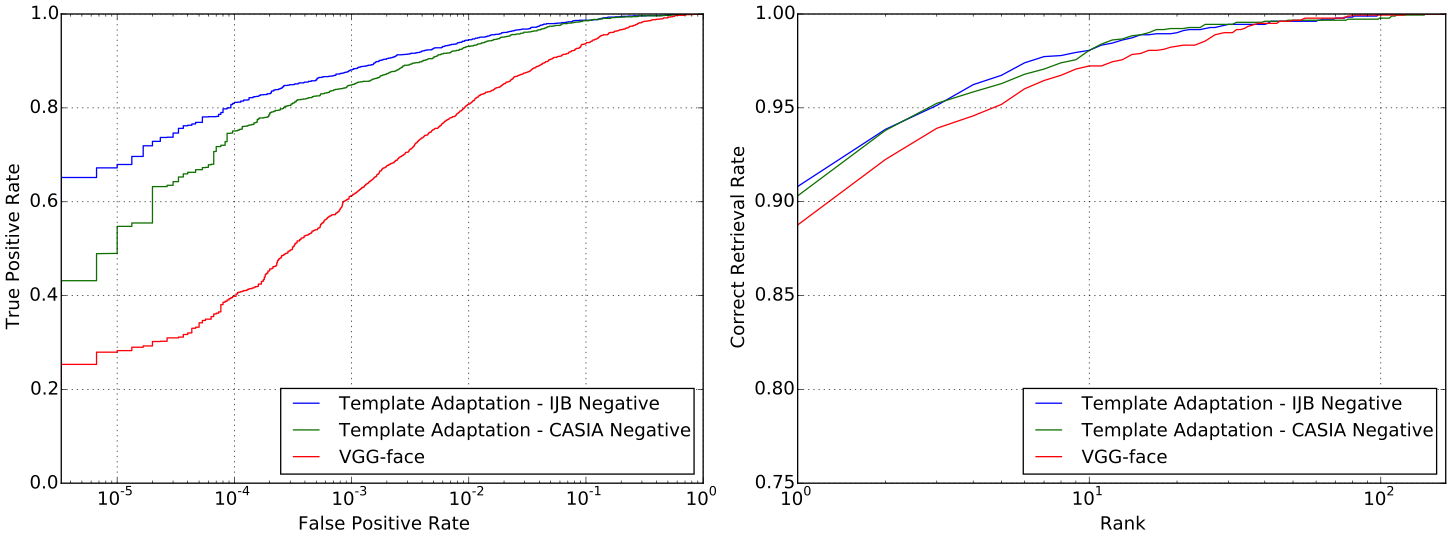} 
\caption{Negative Set Analysis.  We compare the effect of different negative sets for template adaptation.  (top) The best choice is using the other non-mated gallery templates to define the negative set.  (bottom) Experiments with a large unrelated negative set based on CASIA WebFaces results in slightly lowered performance.}
\label{f:negativeset}
\end{centering}
\end{figure*}

%% file: fig_analysis_of_alternatives.tex
\begin{figure*}[!t]
\begin{centering}
\includegraphics[width=\figwidth{}]{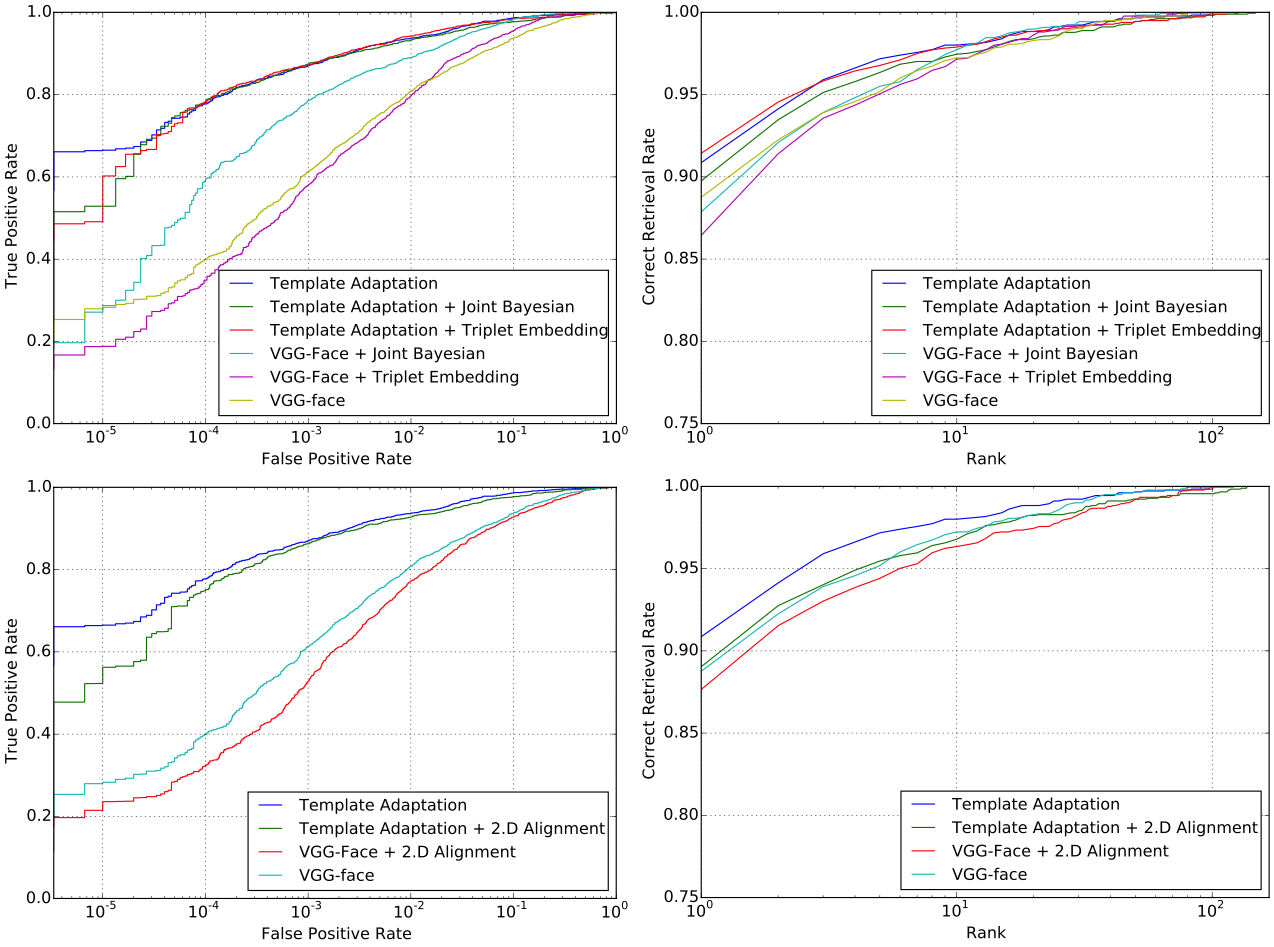} 
\caption{Analysis of Alternatives. We show verification ROC curves (left) and identification CMC curves (right) for IJB-A split-1.  (top) Template adaptation compared with CNN encoding with metric learning using triplet similarity embedding \cite{Parkhi15,Schroff15} or Joint Bayesian embedding \cite{Chen12,Chen15}. (bottom) Template adaptation compared with CNN encoding and 2D alignment \cite{Taigman14,Parkhi15}.  In both cases, template adaptation outperforms all methods, and when combined with metric learning or 2D alignment, generates nearly equivalent performance.  }
\label{f:aoa}
\end{centering}
\end{figure*}

%% file: fig_templatesize.tex
\begin{figure*}[t]
\begin{centering}
\includegraphics[width=\figwidth{}]{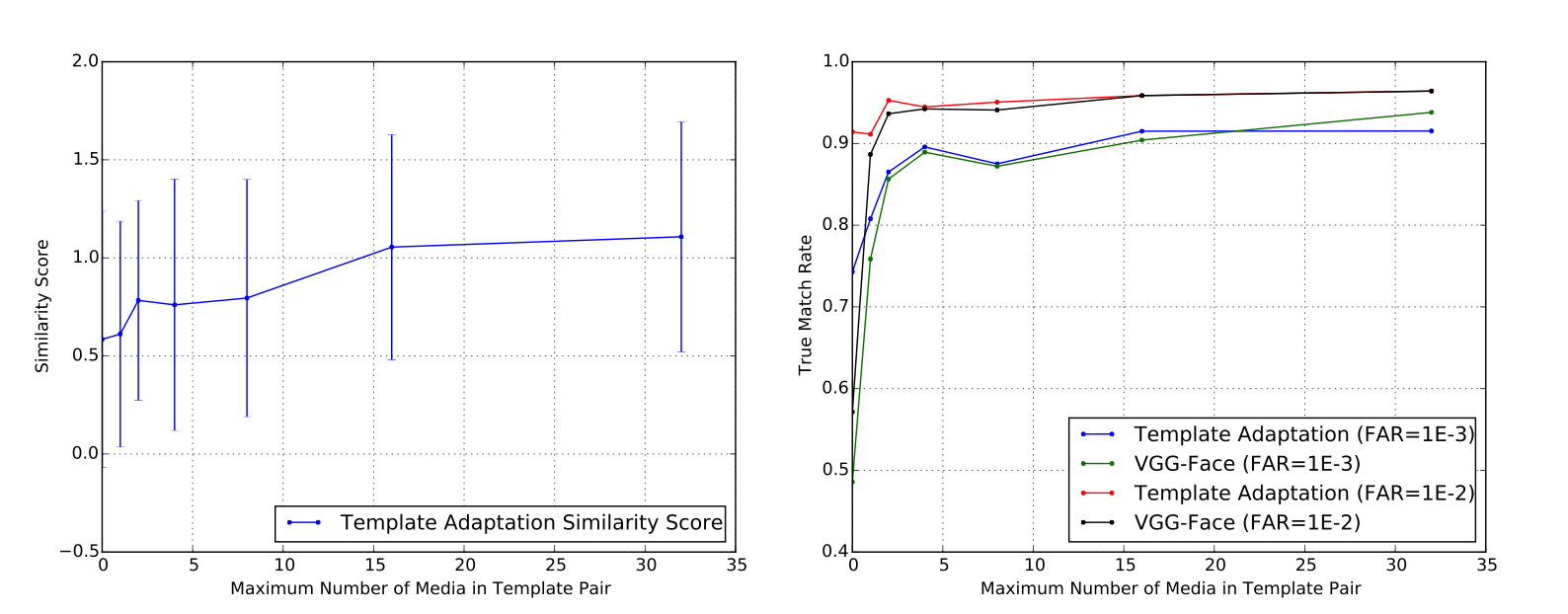} 
\caption{Template size analysis. (left) Similarity score increases as a function of maximum number of media, where the standard deviation is largest when template size is one. (right) True match rate as a function of maximum number of unique images or videos in a template pair, which shows that verification performance levels off at a maximum of {\em three} unique media per template.}
\label{f:template}
\end{centering}
\end{figure*}

%% file: fig_fusion.tex
\begin{figure*}[!t]
\begin{centering}
\includegraphics[width=\figwidth{}]{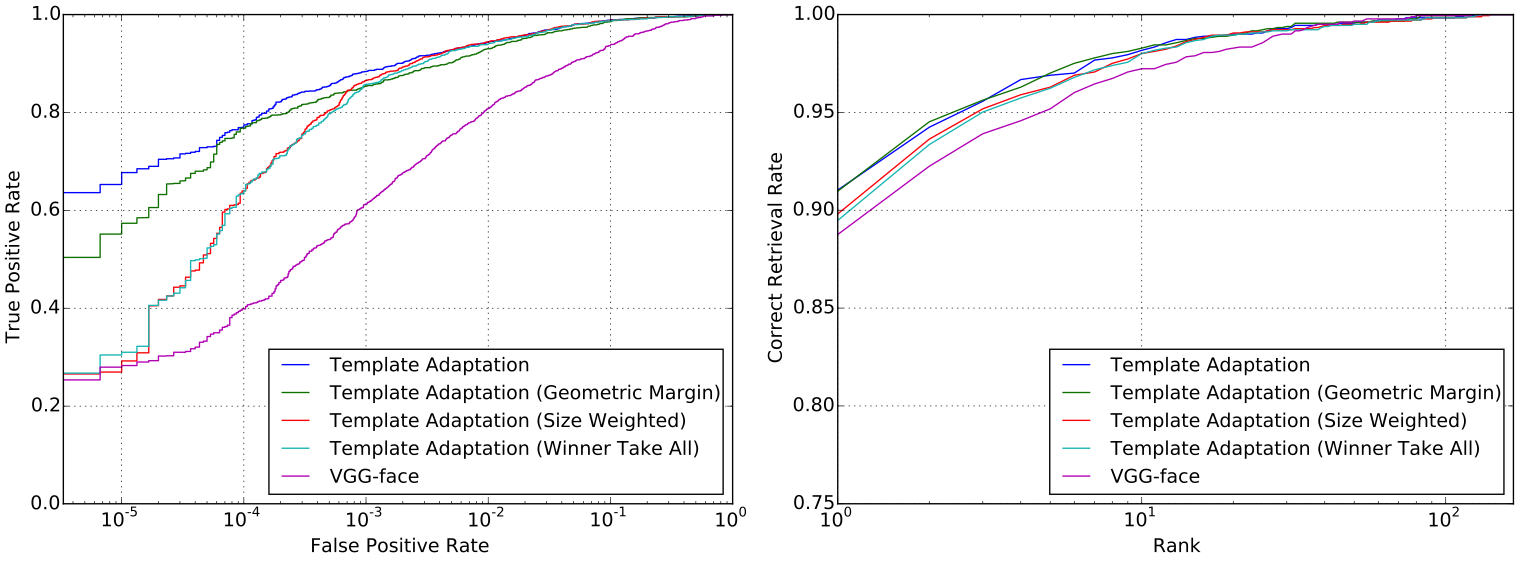} 
\caption{Classifier fusion study.  We compare strategies for linear weighted fusion of classifiers, and results show that an average fusion used by the default template adaptation is best.}  
\label{f:fusion}
\end{centering}
\end{figure*}

%% file: fig_cs2_d21_verification.tex
\begin{figure*}[t]
\begin{centering}
\includegraphics[width=\figwidth{}]{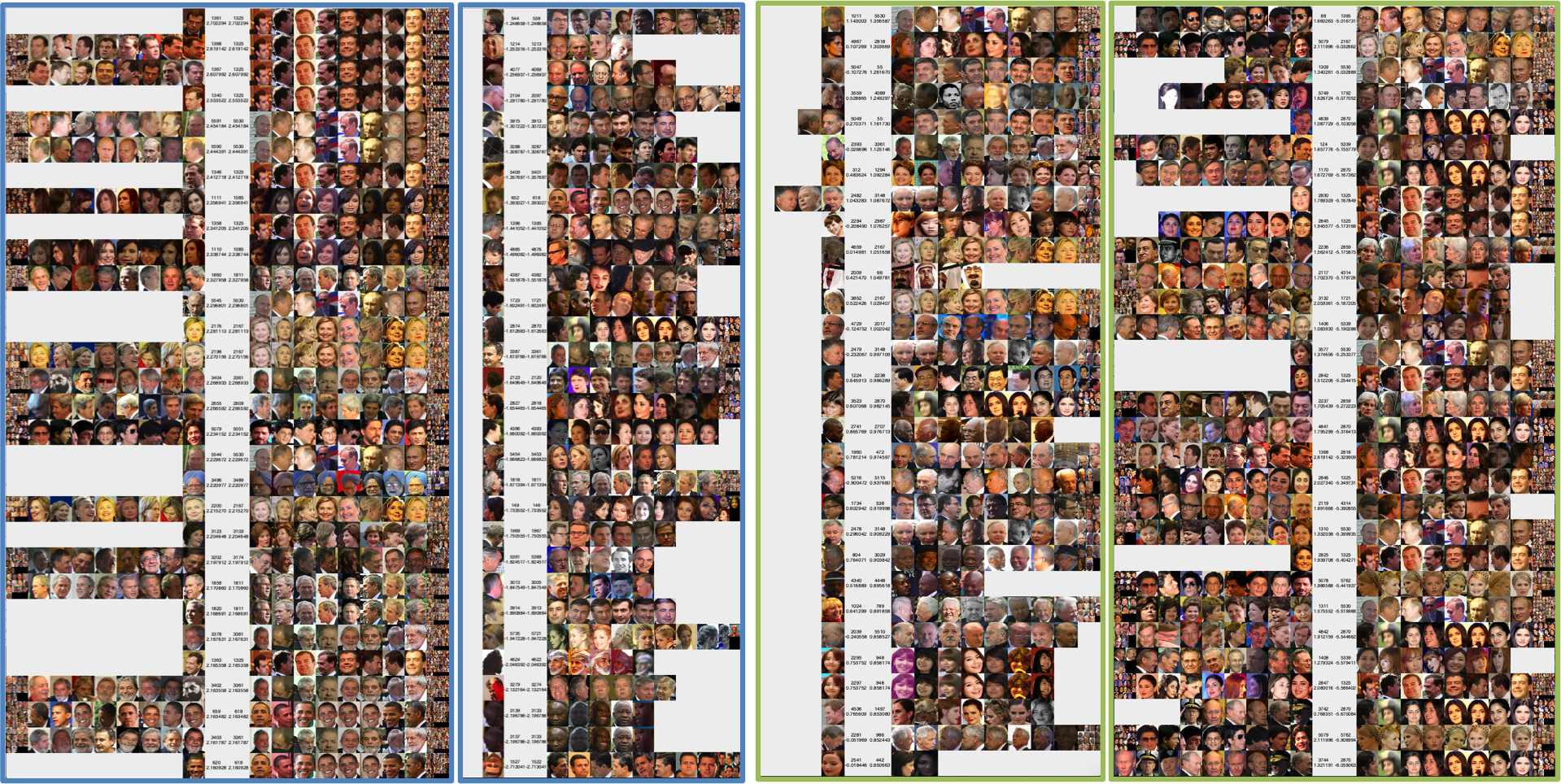} 
\caption{Verification error analysis.  (far left, blue) The best mated verification template pairs, (center left, blue) The worst mated verification template pairs, (center right, green) The worst non-mated verification template pairs (far right, green) The best non-mated verification template pairs.  
}
\label{f:verification_errors}
\end{centering}
\end{figure*}